\documentclass[sigconf]{acmart}
\AtBeginDocument{%
  }

\usepackage{algorithm2e}

\setcopyright{acmlicensed}
\copyrightyear{2026}
\acmYear{2026}
\acmDOI{10.1145/3815598.3815632}
\acmConference[FDG '26]{Foundations of Digital Games}{August 10--13, 2026}{Copenhagen, Denmark}
\acmBooktitle{Foundations of Digital Games (FDG '26), August 10--13, 2026, Copenhagen, Denmark}
\acmISBN{979-8-4007-2495-4/2026/08}

\begin{document}

\title{Finite Automata Extraction: Low-data World Models as Programs from Gameplay Video}

\author{Dave Goel}
\email{dgoel1@ualberta.ca}
\affiliation{
  \institution{Department of Computing Science, Alberta Machine Intelligence Institute (Amii), University of Alberta}
  \city{Edmonton}
  \state{Alberta}
  \country{Canada}
}

\author{Matthew Guzdial}
\email{guzdial@ualberta.ca}
\affiliation{%
  \institution{Department of Computing Science, Alberta Machine Intelligence Institute (Amii), University of Alberta}
  \city{Edmonton}
  \state{Alberta}
  \country{Canada}
}

\author{Anurag Sarkar}
\email{sarkar4@ualberta.ca}
\affiliation{%
  \institution{Department of Computing Science, Alberta Machine Intelligence Institute (Amii), University of Alberta}
  \city{Edmonton}
  \state{Alberta}
  \country{Canada}
}


\begin{abstract}
World models are often neural network-based models that attempt to approximate an entire video game. 
Existing world models are not practical for game developers due to their large scale, lack of accessibility, their prompt-based interfaces, and their nature as black box models where developers cannot access the code. 
In this paper, we propose an approach, Finite Automata Extraction (FAE), that learns a neuro-symbolic world model represented as programs in a novel domain-specific language (DSL): Retro Coder. 
The neuro-symbolic world model structure allows for low data learning, meaning we can learn these models from a single gameplay video.
This model's structure means that developers can directly access code to change the behaviour of the model, and it's low data cost makes it more accessible.
In a comparison to prior neural world model approaches, FAE learns a more precise model of the environment and it learns more general code than prior DSL-based approaches. 
\end{abstract}



\maketitle

\section{Introduction}

Modern video games are primarily made using game engines\cite{engines}. 
Game developers are typically required to have technical depth and breadth in order to use these engines, needing knowledge of aspects such as programming languages, mathematics, physics concepts, AI implementations and so on. 
As a result, the barrier to entry to becoming a game developer can be quite high. 
With the arrival of the third AI summer, there has been a trend to develop tools that can lower this barrier \cite{aitools}. Some industry researchers have recently proposed world models as a tool for game development \cite{gamengen}.

World models are a compressed spatial and temporal learned representation of an environment \cite{worldmodels}. They are predictive models that mimic an environment and approximate future states based on the actions of an agent. 
For example, creating a simplified, playable clone of Pac-Man in a neural network as in NVIDIA's GameGAN \cite{gamengen}.
Researchers typically apply world model learning to reinforcement learning or other automated game playing approaches \cite{worldmodels}. Recently, some industry researchers have moved towards motivating world model learning for game creation \cite{genie2, gamengen}. They claim that world models can act as game engines, and a particular subset of the latent space/world model can be devoted to a single playable game.
Thus game development would become as simple as searching through this latent space of possible games.
However, there are some issues with this approach. These models do not have a game state or memory, and thus have a tendency to hallucinate or lose track of offscreen objects \cite{hallucinations}. This could make it hard for developers to make a cohesive gameplay loop, since the mechanics in the environment would be unpredictable and unreliable. Since these playable world models are all neural networks, they are also a black box to the developer \cite{blackbox}. 
They cannot make changes to the behaviour of the model as they could with traditional game code. Training these models also requires a large amount of data and computation, resources which may not be available to the majority of developers. As a result, the only viable option for developers would be to use pretrained models, which further reduces their autonomy. Such models like Genie 3, can be accessed with a paid subscription \cite{genie3}. Recently, OpenAI shut down SORA, a video generation model, raising concerns around long-term access.

If we had a world model learning approach that better suited the needs of game developers in terms of an editable code representation and only requiring low data, it could be a powerful tool for development. 
This paper presents Finite Automata Extraction (FAE): a novel neuro-symbolic approach to world model learning from a single gameplay video. A neuro-symbolic pipeline uses a combination of neural network architectures and symbolic approaches.
Our pipeline learns a dictionary of sprites in a gameplay video using a neural network and then searches over Retro Coder, a novel Domain Specific Language (DSL) to extract symbolic programs for each of the learned sprites. A DSL is a custom simplified language created for a specific domain.  This approach would give developers access to the underlying code behind the world model. This would remove the black box nature of a traditional world model, allowing developers to directly alter the mechanics of the game and remove unintended unpredictability. Our approach also uses substantially less data and is less computationally intensive than previous world model approaches. We argue that this would make it more accessible to more game developers. As an initial implementation, our pipeline  targets simple grid based 2D games. In this paper we present the idea and initial evidence to support the feasibility of our pipeline.

We apply our approach on two domains, Pac-Man and River Raid and compare our results to GameGAN, a neural world model \cite{GameGan}, and Game Engine Learning, a DSL-based world model \cite{guzdial2017game}.
In this paper, we present the following contributions:
\begin{itemize}
    \item A novel neuro-symbolic approach to learn world models using gameplay video: Finite Automata Extraction.
    \item A novel Domain Specific Language: Retro Coder.
    \item The results of our evaluation comparing to existing neural and DSL-based world models.
\end{itemize}

\section{Background}
In this section we first cover the technical background knowledge needed to understand our work. 

\subsection{Program Synthesis}
Program synthesis refers to the task of generating programs that satisfy user needs \cite{program_synthesis}. Commonly, developers define a search space of programs from which a desired program can be selected. Typically, this is done by selecting or writing a Domain Specific Language, as the search space in a custom DSL is usually much smaller than a Turing complete language. The developers then define a set of constraints or heuristic for the desired program \cite{gulwani2017program}. For instance, they might give a series of input output pairs for the program. The goal of the search algorithm then would be to search for a program that satisfies most, if not all of these examples.
Program synthesis has long been an area of interest for artificial intelligence research \cite{manna1975knowledge,waldinger1969prow,pnueli1989synthesis} with more recent efforts leveraging neural network-based approaches \cite{balog2017deepcoder,devlin2017robustfill,ellis2018learning,ellis2023dreamcoder}. 

\subsection{MarioNette}

MarioNette is a self-supervised neural network model that decomposes video frames into sprites, i.e distinct pieces of image data stored in a game\cite{Marionette}. It is a regularized Variational Auto-encoder that takes in video frames as input and reconstructs them. It does this by learning a sprite dictionary, a bounded set of neural sprites, i.e., sprite embedding learned by Marionette approximating true sprites.
It uses this dictionary to reconstruct the frame. To do this, it employs a grid representation of the frame, where each grid cell contains a sprite from the sprite dictionary. The model has been found successful at this task for the domain of Super Mario Bros.. We leverage Marionette as the basis for the neural network  in our approach. 
By learning sprites, we can avoid the requirements of prior DSL-based world models in having to provide all possible sprites as the developer \cite{guzdial2017game}.

\section{Related Work}

In this section we first cover prior work in program synthesis, given that Finite Automata Extraction (FAE) relies on it to learn appropriate programs in our Domain Specific Language, Retro Coder. We then cover prior work on world models. 

\subsection{Program Synthesis in Games}

Within the specific context of games, program synthesis has been used for generating boss fights \cite{butler2017program}, solving grid-based games \cite{silver2020few}, generating strategies for RTS games \cite{marino2021programmatic} and generating goal-based minigames \cite{davidson2025goals}.
More commonly program synthesis is used for automated control. \citeauthor{gu2024knowpc} use program synthesis for representing policies for agents in the Overcooked environment \cite{gu2024knowpc}. Similarly \citeauthor{eberhardinger2023learning} synthesize programs to explain an agent's decision-making process in the Atari environment \cite{eberhardinger2023learning}. More recent work has also looked at defining a framework for leveraging large language models to assist with synthesizing programs for accomplishing a variety of in-game tasks \cite{eberhardinger2024code}.
The majority of this prior work does not focus on program synthesis for world model learning, we address exceptions in the next subsection.


\subsection{World Models}
\begin{figure*}[ht]
    \centering
    \includegraphics[width=\textwidth]{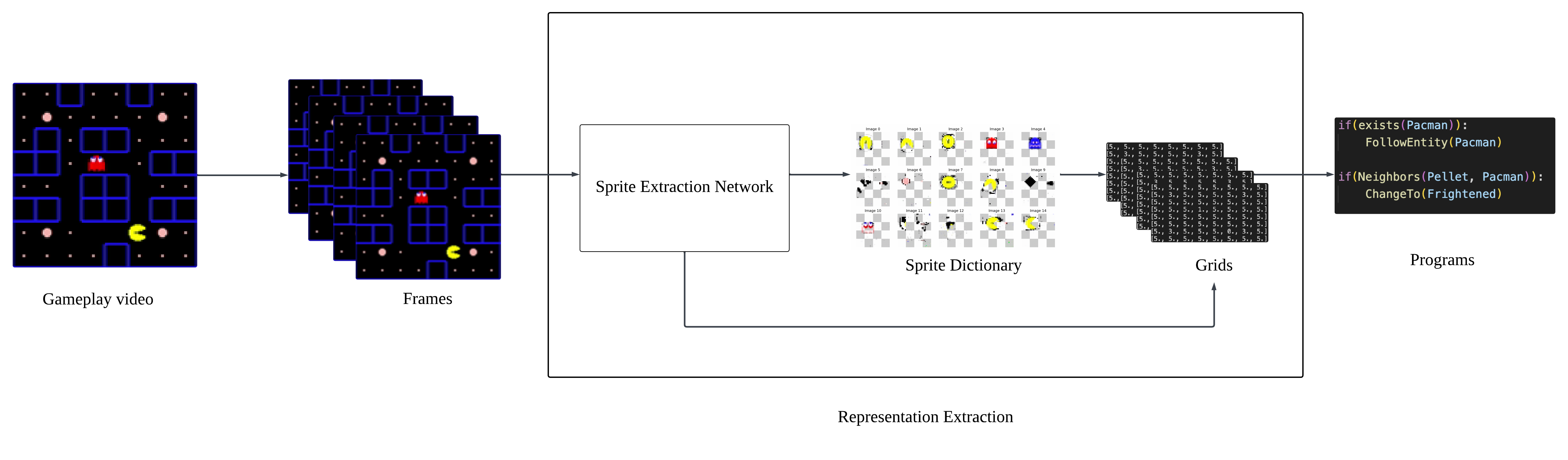}
    \caption{System Overview of Finite Automate Extraction (FAE). We first extract frames from gameplay video, then learn a neural sprite dictionary and a symbolic representation of those frames, and then learn programs for each of the sprites in the sprite dictionary}
    \label{fig:systemOverview}
\end{figure*}

A recent focus has been in training world models that enable real-time interaction via user inputs \cite{menapace2021playable, wu2024ivideogpt}. While some of these works \cite{bruce2024genie, gamengen, genie2, yu2025gamefactory,genie3} are often presented as models capable of mimicking or generating games, they operate by generating video frames conditioned on user input in real-time and lack explicit knowledge of underlying game rules and mechanics.
Due to their size, the majority of the recent world models cannot run in real-time on consumer hardware and are not open source. For these reasons we compare to only GameGAN in our experiments \cite{GameGan}. GameGAN uses a Generative Adveserial Network to predict the next frame, It takes in the previous frame and keyboard actions as its input.

Some prior work has looked at more explicitly modelling of game mechanics such as Game Engine Learning, which employed a Domain Specific Language of if-then rules in a program synthesis-based symbolic learning approach \cite{guzdial2017game}. Our approach differs from this in two ways. Firstly, our pipeline includes learning the sprites, which is not the case with Game Engine learning. Secondly, we use a bottom up approach to learn the programs, as opposed to Game Engine Learning, which uses a top down approach. As a result, our approach learns rules that are more general.
We compare against this approach as well in our experiments as the most similar example of purely symbolic prior work. There are other symbolic approaches, such as learning hybrid automata  \cite{summerville2017mechanics} but they depend on access to an emulator instead of purely gameplay video.

\section{System Overview: Finite Automata Extraction}

In this section, we cover our proposed neurosymbolic approach to learn programs from gameplay video: Finite Automata Extraction (FAE). We give an overview of our pipeline in Figure \ref{fig:systemOverview}. We begin by splitting the gameplay video data into RGB frames. We then train a self-supervised model based on the Marionette architecture \cite{Marionette} on these frames to extract a neural sprite dictionary, a bounded set of sprites. After training the model, we extract a learned symbolic representation of the frames.
We then learn a program for each learned sprite to replicate their behaviour in the gameplay video. We learn each program by searching over Retro Coder, a novel domain-specific language.

\subsection{Representation Extraction}
In this paper, we use entities to refer to unique 2D game objects. Each entity has a Sprite-Class, which dictates it's appearence and behaviour. For example, Blinky, Pellets and Pac-Man would all be different Sprite-Classes.
We require the set of all entities and the positional information of each entity in each frame for the later program learning step of FAE. We settled on a grid representation, where each grid is a matrix of symbolic values, where each value corresponds to a Sprite-Class. This enables us to handle collisions discretely. 
In order to obtain the grid representation, we chose to learn the sprite dictionary consisting of a maximum $l$ number of sprites using a variant of the Marionette architecture \cite{Marionette}. 

We modified the original Marionette network to add an option to specify the background image, which is helpful for domains with a constant background. We changed the model to subtract the background image from both the input and output to ensure the model optimized only the reconstruction of the changing sprites. 
The original Marionette has multiple different loss components, including one for frame reconstruction. 
We added an additional multiplicative weight ($\gamma$) to this loss to further bias the model towards better reconstructions, which incentivizes learned sprite quality. 
We set $\gamma=10$ based on early experimentation.

\begin{figure}[!ht]
    \centering
    \includegraphics[width=0.5\textwidth]{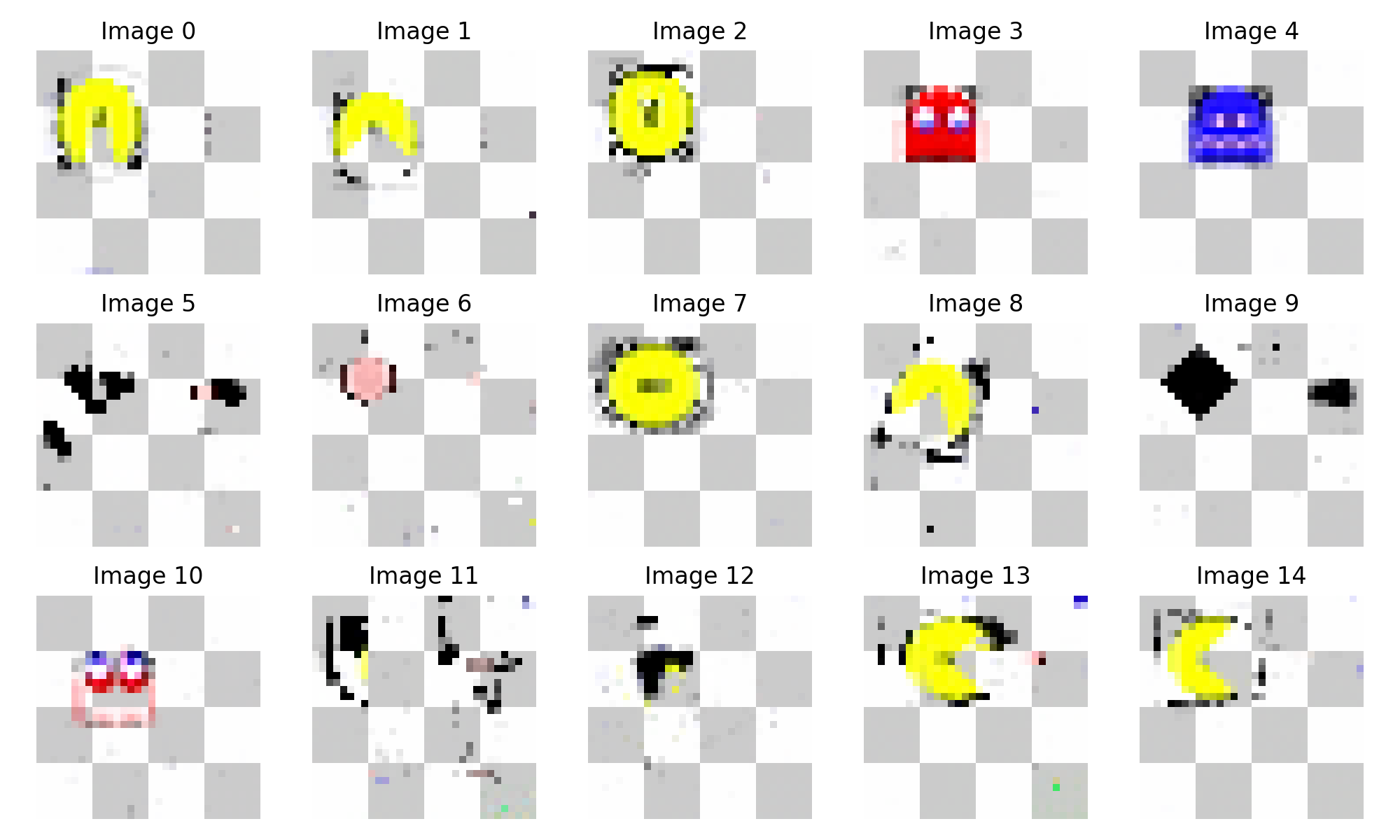}
    \caption{Learned Sprite Dictionary}
    \label{fig:dictionary}
\end{figure}

After training, for all the RGB  frames \{$F_{1},... F_{n}$\}, we extract their grid representations \{$G_{1},... G_{n}$\}. Marionette employs a latent representation of frames in terms of grids of sprite embeddings. Thus we could extract these grids via passing the frames through the model and using the softmax weights from the model to determine the sprite present in each grid cell. 
This results in a $w\times h$ grid where each cell references a learned sprite in the sprite dictionary.  As shown in Figure \ref{fig:dictionary},  the sprite dictionary consists of repeated sprites. For instance, index 0,1,2,7,8,13 and 14 all have sprites that resemble Pac-Man. To resolve this, we group similar looking sprites that imply similar behaviour together into same Sprite-Class. We then modify the $w\times h$ grid so that each cell in the grid refers to the Sprite-Class of the learned sprite.

\subsection{Retro Coder}
For this project we required a DSL that would support learning sprite behaviours. We thus defined our custom Domain Specific Language, Retro Coder. Retro Coder consists of condition functions (if) and action functions (then).  Each if statement has one condition and one action.
Since we designed FAE with 2D games in mind, we focused on common sprite behaviours and conditions for these behaviours. These consist of movement logic, entity transformation logic, and entity collision logic \cite{noah}.
The condition functions are:
\begin{itemize}
    \item $exists\_in\_map(sprite)$ checks if an entity of sprite-class $sprite$ exists in the frame 
    \item $neighboring(sprite)$ checks if the entity of the program itself is next to an entity of sprite-class $sprite$, 
    \item $neighbours(sprite1, sprite2)$ checks if an entity of sprite-class $sprite1$ is next to an entity of sprite-class $sprite2$. Since we are using a grid representation where one entity can exist at one location in a moment of time, we treat neighbouring sprites as collisions. 
\end{itemize}
The action functions are:
\begin{itemize}
    \item $follow\_entity(sprite)$ moves this entity toward the nearest entity of sprite-class $sprite$  
    \item $follow\_direction(direction)$ moves this entity in a direction (Up, down, left, right).
    \item $change\_to\_sprite(sprite)$ changes the entity's sprite-class to $sprite$
    \item $follow\_target\_location(x,y)$, which moves the entity towards to location $x,y$.
\end{itemize}

In the above we note that at times multiple entities of the same sprite-class. may appear on screen.
For example, in Pac-man there are multiple pellets on screen at once.

\subsection{Search Algorithm}

\begin{algorithm}[ht!]
\caption{Pseudocode for Search Algorithm}
\label{tab:algorithm}

$programs \gets [\ ]$\;

\For{$sprite \in sprites$}{
    $program \gets [\ ]$\;

    \For{$batch \in (listofIndicesWhereSpriteExists, batchSize)$}{

        $distanceDecreased \gets True$\;

        \While{$distanceDecreased == True$}{
            $distanceDecreased \gets False$\;

            $neighbours \gets NeighbourPrograms()$\;

            $bestNeighbour \gets program$\;

            \For{$neighbour \in neighbours$}{

                \If{$D(batch, neighbour) < D(batch, bestNeighbour)$}{
                    $bestNeighbour \gets neighbour$\;

                    $distanceDecreased \gets True$\;
                }
            }

            \If{$bestNeighbour \neq [\ ]$}{
                $program \gets bestNeighbour$\;
            }
        }
    }

    $programs.\texttt{append}(program)$\;
}

\end{algorithm}
This section covers the program synthesis step of FAE. The first step is that after we have our sprite-classes, we must categorize the sprite-classes into Learnable and Non-learnable sprites. This might be based on domain knowledge but can also be general at times. For example, the sprite-class that is controlled by the player would be Non-learnable as it does not have a ground truth program.  We then learn a program for all of these hand-identified Learnable sprite-classes with our search algorithm, as seen in \ref{tab:algorithm}. We learn one program at a time, as this greatly reduces the search space. 
For a sprite-class $s$, we first scan \{$G_{1},... G_{n}$\} to get a list of indices \{$I_{1},... I_{s}$\} where $G_{I_{i}}$ contain $s$. We then go through the indices sequentially, for every $i$, compute the distance $D=\sum_{x=i}^{i+b}A.E(G_{I_{x+1}}, G_{I_{x}})$
where $A.E$ is the Absolute Error and $b$ is the batch size. We set $b=3$ based on experimentation. We then search for neighbour programs to minimize $D$. The neighbour programs can either have an additional if-then statement, or remove an if-then statement from the current program.
We consider all possible additions and all possible deletions in the $NeighbourPrograms()$ function.

\section{Evaluation}
Finite Automate Extraction allows us to process a single gameplay video of a 2D game and extract a learned set of sprites and a learned set of programs, one associated with each sprite.
We evaluate our model from two perspectives: the quality of the learned programs, which assesses the backend of the model, and the quality of the world model as an approximation of a game, which assesses the front end. We test our model on two domains, Pac-Man and River Raid. Since this is an initial implementation, and thus primarily targeted at simple grid based 2D games,
both of these meet the criteria while differing in terms of the types of games, with one focused on combat and the other on collecting.

The programs were first learned using a single training video of each game, after which we used a test video for evaluation. To generate the frames from the model, we used 2 approaches: Next Frame and Total Simulation. For Next Frame we use the ground truth frame at each time step $t$ to predict the next frame with our model at time step $t+1$. For Total Simulation, we only use the ground truth frame at time step $t=0$, after which the predicted frame at timestep $t$ is used to predict the frame at timestep $t+1$.
Next Frame allows us to evaluate local coherency while Total Simulation evaluates global coherency of the learned programs.

\subsection{Domains}

\begin{figure*}[ht]
    \centering
    \includegraphics[]{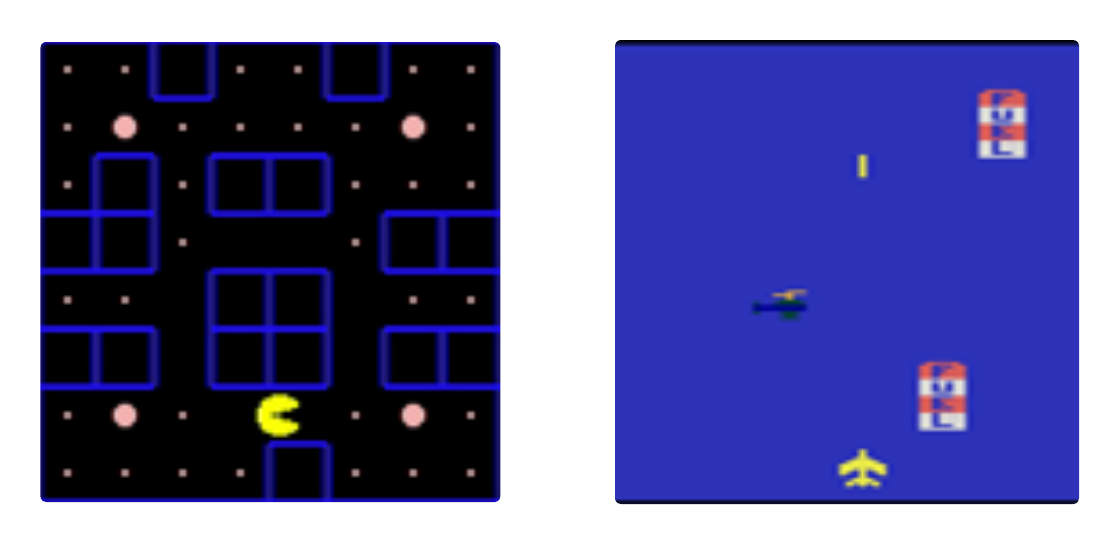}
    \caption{Screenshots of gameplay of both domains, Pac-Man and River Raid.}
    \label{fig:screenshots}
\end{figure*}

Pac-Man is an arena-based  game where the player navigates a level and collect pellets while avoiding ghosts that can kill Pac-Man if they make contact. The game normally features 4 ghosts, Blinky, Inky, Pinky and Clyde. We used a simplified version of the game developed with Unity by removing all the ghosts except Blinky. We also reduced the size of the map from 20x31 to 8x8. 
We recorded the gameplay videos for Pac-Man ourselves. We extracted frames at a rate of four frames per second based on the average human reaction time is 0.25s \cite{reaction}. 

The width $w$ and height $h$ of the learned grid were set to 8, and the maximum number of sprites in the dictionary $l$ was set to 15. We trained the model for 1000 epochs.

River Raid is a scrolling shooter game where the player controls an airplane and has to shoot obstacles. While recording, we cropped the screen to remove background details that were not involved in the gameplay to better focus on the sprites and their behaviours. 
We used the Atari gym environment to record the gameplay videos for River Raid. We used the same hyper parameters from Pac-Man with the exception that $l$ was set to 16.

\subsection{Baselines}

For a neural world model baseline, we used GameGAN \cite{GameGan}. 
We believe that this model is a reasonable representation of current world models. Further the largest of modern world models like GameNGen and Genie 3 are not accessible to the public, and could not be retrained with our resources even if they were. We trained GameGAN on the same Pac-Man and River Raid data we used to train FAE. 
Notably this is far less data than GameGAN had in the original publication. Our goal was to see how well a neural world model could do with this limited amount of data.

We use Game Engine Learning \cite{EngineLearning} as a DSL-based world model baseline. Game Engine Learning is an existing symbolic world model that uses a top down approach, we wanted to see how well our algorithm performs again GEL. We again used the same training data that we used for FAE.

\begin{table*}[ht]
\centering

\begin{minipage}{0.48\textwidth}
\centering
\begin{tabular}{ll}
\hline
\textbf{Sprite} & \textbf{Learned Program} \\ \hline
\textbf{BLINKY} &
  \begin{tabular}[c]{@{}l@{}}
  IF (exists in map(PACMAN)) \\
  THEN follow entity(PACMAN) \\
  IF (neighboring entities(PACMAN, PELLET)) \\
  THEN change to entity(GHOST) \\
  IF (is neighboring(GHOST)) \\
  THEN change to entity(PACMAN)
  \end{tabular} \\
\addlinespace
\textbf{GHOST}&
  \begin{tabular}[c]{@{}l@{}}
  IF (is neighboring(PACMAN)) \\
  THEN change to entity(PACMAN) \\
  IF (is neighboring(PACMAN)) \\
  THEN change to entity(EMPTY)
  \end{tabular} \\
\addlinespace
\textbf{PELLET}&
  \begin{tabular}[c]{@{}l@{}}
  IF (is neighboring(PACMAN)) \\
  THEN change to entity(PACMAN)
  \end{tabular} \\
\addlinespace
\textbf{EYES}&
  \begin{tabular}[c]{@{}l@{}}
  IF (exists in map(PACMAN)) \\
  THEN change to entity(EMPTY)
  \end{tabular} \\ \hline
\end{tabular}
\caption{Programs Generated with Finite Automata Extractor for Pac-Man}
\label{tab:pacman_programs}
\end{minipage}
\hfill
\begin{minipage}{0.48\textwidth}
\centering
\begin{tabular}{ll}
\hline
\textbf{Sprite} & \textbf{Learned Program} \\ \hline
\textbf{BLINKY} &
  \begin{tabular}[c]{@{}l@{}}
  IF (exists in map(PACMAN)) \\
  THEN follow entity(PACMAN) \\
  IF (neighboring entities(PACMAN, PELLET)) \\
  THEN change to entity(GHOST)
  \end{tabular} \\
\addlinespace
\textbf{GHOST}&
  \begin{tabular}[c]{@{}l@{}}
  IF (is neighboring(PACMAN)) \\
  THEN change to entity(EYES)
  \end{tabular} \\
\addlinespace
\textbf{PELLET}&
  \begin{tabular}[c]{@{}l@{}}
  IF (is neighboring(PACMAN)) \\
  THEN change to entity(PACMAN)
  \end{tabular} \\
\addlinespace
\textbf{EYES}&
  \begin{tabular}[c]{@{}l@{}}
  IF (exists in map(PACMAN)) \\
  THEN follow target location(4,4)
  \end{tabular} \\ \hline
\end{tabular}
\caption{Ground Truth Programs}
\label{tab:pacman_ground_truth_programs}
\end{minipage}

\end{table*}

\begin{table*}[ht]
\centering

\begin{minipage}{0.48\textwidth}
\centering
\begin{tabular}{ll}
\hline
\textbf{Sprite} & \textbf{Learned Program} \\ \hline
\textbf{PELLET} &
  \begin{tabular}[c]{@{}l@{}}
  IF (exists in map(PLANE)) \\
  THEN follow direction(UP) \\
  IF (exists in position(4,1)) \\
  THEN change to entity(EMPTY) \\
  IF (exists in position(4,3)) \\
  THEN change to entity(EMPTY) \\
  IF (exists in position(3,1)) \\
  THEN change to entity(EMPTY)
  \end{tabular} \\
\addlinespace
\textbf{BLACKBOAT} & \\
\addlinespace
\textbf{FUEL} &
  \begin{tabular}[c]{@{}l@{}}
  IF (exists in position(7,7)) \\
  THEN follow target(6,7) \\
  IF (is neighboring(PELLET)) \\
  THEN change to entity(EMPTY)
  \end{tabular} \\
\addlinespace
\textbf{GREENBOAT} &
  \begin{tabular}[c]{@{}l@{}}
  IF (exists in map(PLANE)) \\
  THEN follow target(1,2)
  \end{tabular} \\ \hline
\end{tabular}
\caption{Programs Generated with Finite Automata Extractor for River Raid}
\label{tab:riverraidprograms}
\end{minipage}
\hfill
\begin{minipage}{0.48\textwidth}
\centering
\begin{tabular}{ll}
\hline
\textbf{Sprite} & \textbf{Learned Program} \\ \hline
\textbf{PELLET} &
  \begin{tabular}[c]{@{}l@{}}
  IF (exists in map(PLANE)) \\
  THEN follow direction(UP) \\
  IF (is neighboring(BLACKBOAT)) \\
  THEN change to entity(EMPTY) \\
  IF (is neighboring(GREENBOAT)) \\
  THEN change to entity(EMPTY)
  \end{tabular} \\
\addlinespace
\textbf{BLACKBOAT}&
  \begin{tabular}[c]{@{}l@{}}
  IF (exists in map(PLANE)) \\
  THEN follow direction(DOWN) \\
  IF (is neighboring(PELLET)) \\
  THEN change to entity(EMPTY)
  \end{tabular} \\
\addlinespace
\textbf{FUEL}&
  \begin{tabular}[c]{@{}l@{}}
  IF (exists in map(PLANE)) \\
  THEN follow direction(DOWN) \\
  IF (is neighboring(PELLET)) \\
  THEN change to entity(EMPTY)
  \end{tabular} \\
\addlinespace
\textbf{GREENBOAT}&
  \begin{tabular}[c]{@{}l@{}}
  IF (exists in map(PLANE)) \\
  THEN follow direction(DOWN) \\
  IF (is neighboring(PELLET)) \\
  THEN change to entity(EMPTY)
  \end{tabular} \\ \hline
\end{tabular}
\caption{Ground Truth Programs for River Raid}
\label{tab:riverraidgroundtruthprograms}
\end{minipage}

\end{table*}

\begin{table*}[ht]
\centering
\begin{tabular}{lclclc}
\toprule
\textbf{Method} & \textbf{FID(train)}&\textbf{FID(test)}& \textbf{ Error(train)}&\textbf{ Error(test)}& \textbf{Conditions} \\
\cmidrule(r){1-1} \cmidrule(l){2-5}
GameGAN& 10.17&10.46& 35.31&40.64& -            \\
Game Engine Learning       & 0.25&0.39& 10.46&6.91& 73±22.55     \\
Finite Automata Extraction & 0.23&0.28& 10.60&7.06& \textbf{1.75±0.83} \\
\bottomrule
\end{tabular}
\caption{Comparison of methods on FID, Predicted Error and average number of conditions for Pac-Man, with the Next Frame approach.}
\label{tab:pacman_results}
\end{table*}

\begin{table*}[ht]
\centering
\begin{tabular}{lclclc}
\toprule
\textbf{Method} & \textbf{FID(train)}&\textbf{FID(test)}& \textbf{ Error(train)}&\textbf{ Error(test)}& \textbf{Conditions} \\
\cmidrule(r){1-1} \cmidrule(l){2-5}
GameGAN& 10.17&10.46& 35.31&40.64& -            \\
Game Engine Learning       & 0.19&0.23& 10.87&7.16& 73±22.55     \\
Finite Automata Extraction & 0.28&0.29& 10.37&7.56& \textbf{1.75±0.83} \\
\bottomrule
\end{tabular}
\caption{Comparison of methods on FID, Predicted Error and average number of conditions for Pac-Man, with the Total Simulation approach.}
\label{tab:pacman_results2}
\end{table*}

\begin{table*}[ht!]
\centering
\begin{tabular}{lclclc}
\toprule
\textbf{Method} & \textbf{FID(train)}&\textbf{FID(test)}& \textbf{ Error(train)}&\textbf{ Error(test)}& \textbf{Conditions} \\
\cmidrule(r){1-1} \cmidrule(l){2-5}
GameGAN& 58.62&58.81& 143.62&144.2& -            \\
Game Engine Learning       & 0.37&0.32& 82.42&82.85& 125.42±19.58\\
Finite Automata Extraction & 0.29&0.25& 82.60&83.03& \textbf{2.00±1.87}\\
\bottomrule
\end{tabular}
\caption{Comparison of methods on FID, Predicted Error and average number of conditions for River Raid, with the Next Frame approach.}
\label{tab:riverraid_results}
\end{table*}

\begin{table*}[ht!]
\centering
\begin{tabular}{lclclc}
\toprule
\textbf{Method} & \textbf{FID(train)}&\textbf{FID(test)}& \textbf{ Error(train)}&\textbf{ Error(test)}& \textbf{Conditions} \\
\cmidrule(r){1-1} \cmidrule(l){2-5}
GameGAN& 58.62&58.81& 143.62&144.2& -            \\
Game Engine Learning       & 0.17&0.14& 83.44&84.14& 125.42±19.58\\
Finite Automata Extraction & 0.14&0.13& 83.58&84.04& \textbf{2.00±1.87}\\
\bottomrule
\end{tabular}
\caption{Comparison of methods on FID, Predicted Error and average number of conditions for River raid, with the Total Simulation approach.}
\label{tab:riverraid_results2}
\end{table*}

\subsection{Metrics}

For a reconstruction quality metric we used \textbf{FID}, which is a standard metric for video generation \cite{metrics}. This allows us to capture a notion of the perceptual quality of the learned world models. In addition to using this established metric, we created another metric, \textbf{Prediction Error} $\sum_{i=0}^{l}MAE(T_{i+1}-T_{i},M_{i+1}-T_{i})$  to assess the distance between the ground truth set of frames {$T_{1},... T_{l}$\} and generated set of frames {$M_{1},... M_{l}$\}. 
This allowed us to capture a notion of precision over the sprites instead of over pixels.

It is difficult to assess the quality of learned programs. 
We note that the above metrics do indirectly capture some notion of the functional quality of the learned programs, especially \textbf{Prediction Error}. Outside of this, we also report the number of conditions, as a way to roughly approximate the generalizability of the learned programs. We only report this for FAE and Game Engine Learning, as GameGAN learns no explicit programs.


\section{Results}

In this section we first cover the learned programs for the Pac-Man and River Raid domains. Part of the benefits of FAE is that we can easily extract the learned programs and present them in a readable format. This also allows us to discuss some particulars of the learned programs that may be non-obvious from the metrics-based analysis. We then cover our metric results comparing the performance of FAE against the baselines within both domains. 

\subsection{Learned Programs}

Using our variant of Marionette, FAE was able to learn six distinct sprite-classes for Pac-Man: PACMAN, BLINKY, GHOST, EYES, PELLET, EMPTY. 
PACMAN indicates the player, BLINKY indicates the red ghost sprite, GHOST indicates the blue ghost sprite that BLINKY turns into when PACMAN eats a PELLET, EYES indicates the eye sprite a GHOST turns into when eaten, and EMPTY represents empty space. 
Notably these names are given only for reporting purposes and do not exist in the system.
FAE represents these entities are indices in the learned sprite dictionary.
We could not learn a sprite-class for the normal pellets due to their size, which is a limitation of Marionette. 
We include a human-readable version of the learned programs in Table \ref{tab:pacman_programs} and Table \ref{tab:riverraidprograms} using these entity names.
The DSL encodes If and Then statements implicitly, as such we prepended the ``IF'' and ``THEN'' statements. 
However, the simplicity of this translation and the associated programs themselves indicates a strong potential for explainability. 

We also included the closest ground truth programs for the two domains in Table \ref{tab:pacman_ground_truth_programs} and Table \ref{tab:riverraidgroundtruthprograms}. Do note that there were certain behaviours that could not be captured in our DSL and therefore the programs. We cover these behaviours later in this section.

For Pac-Man, we identified PACMAN and EMPTY as Non-learnable sprite-classes, PACMAN due to player control and BLANK due to a lack of dynamic information.

Even the ground truth program does not fully capture the correct behaviour for GHOST, indicating an issue with our DSL. In the actual game, GHOST moves in random directions. However, the current version of our DSL cannot capture random behaviour. 
As such this was not possible for FAE to learn. 
FAE successfully learned that BLINKY should always follow PACMAN, which can be seen in the first condition of BLINKY. It also predicts that BLINKY should change into GHOST when PACMAN collides or eats a PELLET. The third learned condition in BLINKY never gets satisfied in the game as the GHOST and BLINKY sprites are never together in a frame. This condition was learned due to the grids not being accurate all the time. Occasionally, MarioNette would hallucinate and return a grid with some sprites that do not actually exist in the true frame.
The model learned that PELLET should change into PACMAN once it has been eaten by PACMAN.
This may seem odd, but has the desired effect, a PELLET sprites should disappear when eaten by PACMAN, thus the PELLET turning into the PACMAN that ate it has the same effect.
The model could not replicate the ground truth behaviour for EYES. This is because there were not enough frames in the training video to capture it's behaviour.
The GHOST does learn about being eaten by PACMAN, however, instead of turning into EYES like in the original game, it just disappears. 
These results show that our system has learned two of the four programs correctly. It should be noted that these two sprite-classes occupy considerably more screen time than the other two.
We think this is a reasonable performance from a single gameplay example and could meaningfully support many game development tasks. We provide an example in the Case Study section.

For River Raid, we also learned six sprite-classes, PLANE, FUEL, GREENBOAT, BLACKBOAT and PELLET. PLANE indicates the triangular sprite that the player controls, FUEL indicates the can sprite, GREENBOAT and BLACKBOAT represents the green and black boat sprites respectively. The translated programs for this domain can be seen in \ref{tab:riverraidprograms}.  Like Pac-Man, we identified PLANE and EMPTY as Non-learnable sprite-classes.

\begin{figure*}[ht]
    \centering
    \includegraphics[]{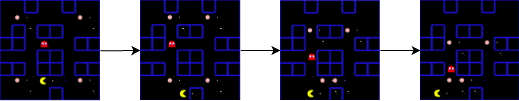}
    \caption{Screenshots of gameplay of Pellet Chaser}
    \label{Pellet Chaser}
\end{figure*}

While the ground truth programs for the game are accurate, there is one mechanic that they cannot cover. In the original game, BLACKBOAT, FUEL and GREENBOAT get randomly spawned on the top row. However, our current DSL cannot capture this behaviour.
The model successfully learned that PELLET should always be moving up. However, it also learned to change to EMPTY for arbitrary positions. The model did not learn any code for BLACKBOAT, which we believe was because the sprite-class appeared rarely in the training video, much like EYES in Pac-Man. FUEL is an interesting case since the size of the sprite is much bigger. As a result, it alternated between occupying two and three grid cells for each frame, depending on its position. This led to the algorithm not being able to learn it's movement behaviour. However, it did learn the collision behaviour of the original game as it changes to empty upon colliding with PELLET. The behaviour for GREENBOAT could not be replicated, which we believe was due to the same reason as BLACKBOAT. These results are less favourable than those of Pac-Man, and demonstrate the importance of DSL choices, since we assumed all sprites would occupy a single grid cell in our DSL.

\subsection{Comparison with Baselines}


In this subsection we discuss the results of our metrics-based comparison with our two baselines.
We first go over the results of the metrics for Pac-Man. 
The results for the two variations can be seen in Tables  \ref{tab:pacman_results} and \ref{tab:pacman_results2}. These show the FID, Predicted Error, and average number of conditions across the models. Since GameGan does not produce any code, we cannot measure the average number of conditions.  
While running the Game Engine Learning System, the algorithm failed to converge due to not being able to make a perfect prediction and so we take the best performing program which follows the practices from prior work \cite{EngineLearning}. 
The average number of conditions were fewer for Finite Automata Extraction than Game Engine Learning (GEL). This shows that the code extracted by our system was more general as fewer conditions would lead to more cases of learned code running. The difference in the number of conditions is based on the way the different algorithms work, GEL starts with the maximum number of conditions then reduces whereas we add conditions as needed. 

We trained GameGAN to until convergence. Due to its neural network nature and the low amount of data, it had the worst performance. 

For the Next Frame approach, across the baselines, FAE had the lowest FID for both the train and test video, though the results are comparable with GEL. This is due to the fact that GameGAN was unable to learn a non-random model from the amount of available training data. 
Further, while GEL may appear to have similar performance this is due to it only predicting the passed in prior frame exactly. This indicates GEL did not learn transferrable information to this test video from the training video. However, predicting the prior frame is a strong baseline itself as established by prior work \cite{EngineLearning}.  For the same reason, the Predictive Error for GEL and FAE is again comparable.

For the Total Simulation approach, GEL and FAE's results were again comparable, although GEL had a lower FID for both test and train. 
This is because GEL did not make any changes to the frame, so every frame output from the model consists of the same image. On the other hand, since FAE was not able to learn the behaviour of GHOST and EYES sprites correctly, the image sequence completely diverges as soon as one of the two sprites appear in the ground truth video. The Predicted Errors are inconclusive since they are lower for FAE for the train video and lower for GEL for the test video.

For River Raid, the FID and Predicted Error results for GameGAN were again the worst due to model's inability to learn a non-random model. The number of conditions were also again less for FAE than GameGAN. Notably, while the FID scores for GEL and FAE were again comparable, FAE had lower FID for both the approaches for both the train and test videos. 

In the original River Raid game, fuel and boats are spawned offscreen randomly. This was not possible for our current DSL since we would need to model the area outside of the game screen, which is currently not possible in the current version of Retro Coder. 

\section{Case Study}
In this section, we show a use case for how game developers could use our pipeline. We take the Pac-man and River Raid world model programs that we have learned, and then experiment with variations of them by altering the code and the sprites. We first make changes to the learned programs so that they resemble the ground truth programs, we assume the developer would have made changes to correct the issues from the learned programs, additionally we think these changes are small enough to be reasonable.
\begin{figure*}[ht]
    \centering
    \includegraphics[]{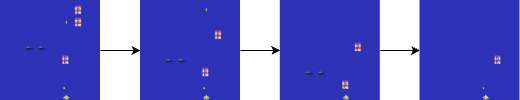}
    \caption{Screenshots of gameplay of River Raid Power Up}
    \label{River raid}
\end{figure*}
\subsection{Example 1: Pellet Chaser}
We changed the code in Pac-Man for PELLET so that it always chases PACMAN. We did this by changing only the if statement in PELLET. This changed the dynamics of the game quite a bit, as the pellets now all seek Pac-Man as seen in Figure \ref{tab:algorithm}. This changes the gameplay from collecting into just avoiding the ghost for long enough until the pellets can catch up with Pac-Man. While we don't think this is a particularly good game on its own, we do find it interesting, and it's the kind of simple thought experiment that FAE could afford to game developers.

\begin{table}[ht!]
\begin{tabular}{ll}
\hline
\textbf{Sprite} & \textbf{Learned Program}                                                                               \\ \hline
\textbf{BLINKY} &
  \begin{tabular}[c]{@{}l@{}}IF (exists in map(PACMAN)) \\ THEN follow entity(PACMAN) \\ IF (neighboring entities(PACMAN, PELLET)) \\ THEN change to entity(GHOST)\end{tabular} \\
\addlinespace
\textbf{GHOST}& \begin{tabular}[c]{@{}l@{}}IF (is neighboring(PACMAN)) \\ THEN change to entity(EYES)\end{tabular}     \\
\addlinespace
\textbf{PELLET}& \begin{tabular}[c]{@{}l@{}}IF (exists in map(PACMAN)) \\ THEN follow entity(PACMAN)\end{tabular}       \\
\addlinespace
\textbf{EYES}& \begin{tabular}[c]{@{}l@{}}IF (exists in map(PACMAN)) \\ THEN follow target location(4,4)\end{tabular} \\ \hline
\end{tabular}
\caption{Program for Pellet Chaser}
\label{tab:pacman_pellet_chaser}
\end{table}

\subsection{Example 2: River Raid Power Up}

We changed the code in River Raid so that the boats get destroyed whenever PLANE touches FUEL. We did this by adding an if statement to both the boats, the condition being that if PLANE and FUEL are neighbouring each other, the boats change to EMPTY. While this maintains the core loop of the game, it changes the dynamics substantially as the player now has the option to collide with FUEL to destroy the obstacles essentially giving the player an area-of-effect (AOE) attack that did not exist in the original game. Again, this is likely not a better game than River Raid, but it allows for an interesting and fast exploration of a variation of the game.
\begin{table}[ht!]
\begin{tabular}{ll}
\hline
\textbf{Sprite} &
  \textbf{Learned Program} \\ \hline
\addlinespace
\textbf{PELLET}&
  \begin{tabular}[c]{@{}l@{}}IF (exists in map(PLANE)) \\ THEN follow direction(UP) \\ IF (is neighboring(BLACKBOAT))\\ THEN change to entity(EMPTY)\\ IF (is neighboring(GREENBOAT))\\ THEN change to entity(EMPTY)\end{tabular} \\
\addlinespace
\textbf{BLACKBOAT}&
  \begin{tabular}[c]{@{}l@{}}IF (exists in map(PLANE)) \\ THEN follow direction(DOWN)\\ IF (is neighboring(PELLET)) \\ THEN change to entity(EMPTY)\\ IF (neighboring entites(PLANE, FUEL)) \\ THEN change to entity(EMPTY)\end{tabular} \\
\addlinespace
\textbf{FUEL}&
  \begin{tabular}[c]{@{}l@{}}IF (exists in map(PLANE)) \\ THEN follow direction(DOWN)\\ IF (is neighboring(PELLET)) \\ THEN change to entity(EMPTY)\end{tabular} \\
\addlinespace
\textbf{GREENBOAT}&
  \begin{tabular}[c]{@{}l@{}}IF (exists in map(PLANE)) \\ THEN follow direction(DOWN)\\ IF (is neighboring(PELLET)) \\ THEN change to entity(EMPTY)\\ IF (neighboring entites(PLANE, FUEL)) \\ THEN change to entity(EMPTY)\end{tabular} \\ \hline
\end{tabular}
\caption{Program for Pellet Chaser}
\label{tab:riverraid_mod}
\end{table}

\section{Limitations}
In this paper, we proposed a framework to learn code from a gameplay video. Our results indicate that we can learn more general programs than a prior DSL-based world model, and can outperform a neural network world model on this low amount of data. However, there are still some limitations to our existing approach. In our current set up, the grid representations that were extracted from the Marionette variant occasionally return inaccurate frames. For example, occasionally a sprite can be between two grid cells. In these instances, our grid representation hallucinated that there was the same sprite in both of the cells. This caused the model to learn conditions that were not true. Perhaps in the future, we can tweak the Marionette Architecture to take this into account. Another aspect that the current representation does not handle is the rotation of the sprites. The sprites currently are rendered in only one rotation. 
As noted in the case of River Raid, the grid based representation in its current state also does not handle sprites with different sizes well.

In its current state, Retro Coder also has some limitations. Since we learn programs for individual sprites, we cannot create a central manager or control program to handle sprite spawning. Additionally, the DSL does not take time into account, and can thus not make behaviours that are time dependent. For example, a certain sprite changing into a different sprite every $x$ seconds. The DSL also cannot replicate randomized behaviour.

\section{Conclusion}
In this paper, we proposed a neuro-symbolic approach to learn a world model as code from gameplay video: Finite Automata Extraction (FAE). We did this by training a model to convert the video frames into a symbolic grid representation. Additionally we defined a novel DSL, Retro Coder as the search space of the world model programs. FAE uses lower data compared to current neural network world models. We tested our pipeline in two domains, Pac-Man and River Raid. Our model was able to learn accurate code for some of the sprites and showed equivalent or better results compared to the existing baselines.

\begin{acks}
This work was funded by the Canada CIFAR AI Chairs Program. We acknowledge the support of the Alberta Machine Intelligence Institute (Amii). 
\end{acks}

\bibliographystyle{ACM-Reference-Format}
\bibliography{aaai25}
\end{document}